\pdfoutput=1

\documentclass[11pt]{article}

\usepackage[preprint]{acl}

\usepackage{times}
\usepackage{latexsym}

\usepackage{graphicx}
\usepackage{wrapfig}
\usepackage{amssymb}
\usepackage{amsmath}
\usepackage{bm}
\usepackage{caption}
\usepackage{multirow}
\usepackage{mathrsfs}
\usepackage{mathtools}
\usepackage{algorithm}
\usepackage{algpseudocode}
\usepackage{soul}
\usepackage{booktabs}

\usepackage[T1]{fontenc}

\usepackage[utf8]{inputenc}

\usepackage{microtype}

\usepackage{inconsolata}

\usepackage{graphicx}

\title{Hierarchical Graph Topic Modeling with Topic Tree-based Transformer}

\author{
 \textbf{Delvin Ce Zhang\textsuperscript{1}},
 \textbf{Menglin Yang\textsuperscript{2}},
 \textbf{Xiaobao Wu\textsuperscript{3}},
 \textbf{Jiasheng Zhang\textsuperscript{4}},
 \textbf{Hady W. Lauw\textsuperscript{5}},
\\
\\
 \textsuperscript{1}The Pennsylvania State University,
 \textsuperscript{2}Hong Kong University of Science and Technology (Guangzhou),\\
 \textsuperscript{3}Nanyang Technological University,
 \textsuperscript{4}University of Electronic Science and Technology of China,\\
 \textsuperscript{5}Singapore Management University
\\
\\
\textsuperscript{1}\texttt{delvincezhang@gmail.com},
\textsuperscript{2}\texttt{menglinyang@hkust-gz.edu.cn},\\
\textsuperscript{3}\texttt{xiaobao002@e.ntu.edu.sg},
\textsuperscript{4}\texttt{zjss12358@std.uestc.edu.cn},
\textsuperscript{5}\texttt{hadywlauw@smu.edu.sg},
}

\begin{document}
\maketitle
\begin{abstract}
Textual documents are commonly connected in a hierarchical graph structure where a central document links to others with an exponentially growing connectivity. Though Hyperbolic Graph Neural Networks (HGNNs) excel at capturing such \emph{graph hierarchy}, they cannot model the rich textual semantics within documents. Moreover, text contents in documents usually discuss topics of different specificity. Hierarchical Topic Models (HTMs) discover such latent \emph{topic hierarchy} within text corpora. However, most of them focus on the textual content within documents, and ignore the graph adjacency across interlinked documents.
We thus propose a Hierarchical Graph Topic Modeling Transformer to integrate both topic hierarchy within documents and graph hierarchy across documents into a unified Transformer. Specifically, to incorporate topic hierarchy within documents, we design a topic tree and infer a \emph{hierarchical tree embedding} for hierarchical topic modeling. To preserve both topic and graph hierarchies, we design our model in hyperbolic space and propose \emph{Hyperbolic Doubly Recurrent Neural Network}, which models ancestral and fraternal tree structure. Both hierarchies are inserted into each Transformer layer to learn unified representations. Both supervised and unsupervised experiments verify the effectiveness of our model. 
\end{abstract}

\section{Introduction}

\begin{figure}
	\centering
	\includegraphics[width=1\linewidth]{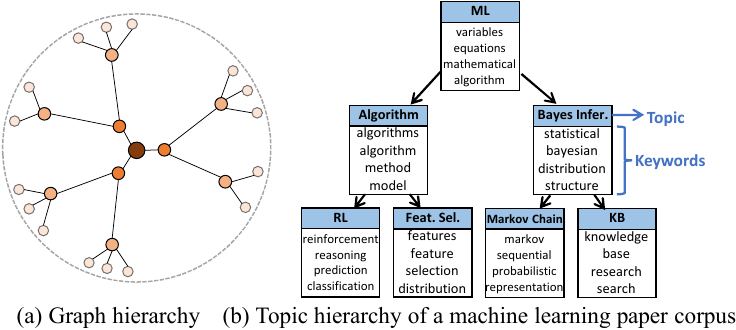}
	\vspace{-0.4cm}
	\caption{(a) Graph hierarchy, (b) topic hierarchy.}
	\label{fig:illustration}
	\vspace{-0.5cm}
\end{figure}

Documents are usually linked as a graph, e.g., papers cited in a citation graph; news articles linked in a hyperlink graph. Such graph usually exhibits a hierarchical structure:
a central document links to others with an exponentially growing connectivity (Fig. \ref{fig:illustration}(a)). For example, an academic paper is extended by follow-up works, which are further cited by others; a news article reporting an event is traced by others with subsequent events. Hyperbolic Graph Neural Networks (HGNNs) \cite{hgcn} capture such \emph{graph hierarchy}. However, when modeling documents, we usually assume latent topics \cite{lda} and model contextualized semantics \cite{transformer}. HGNNs are not designed to capture latent topics or contextualized semantics, leading to inferior document embeddings. Text indicates how documents relate to each other in the latent topic space, and modeling it could capture semantic similarities. 

Moreover, documents usually discuss topics of different specificity. For instance, some news report the overall Olympic Games, while others focus on specific sports; survey papers summarize a broad area, while regular papers deal with specific problems (Fig. \ref{fig:illustration}(b)). Though topic models \cite{wu2023effective,wu2024fastopic,wu2024survey} capture text semantics, most treat all documents equally and infer flat document representations. They fail to explore \emph{topic hierarchy} to differentiate semantic specificity of documents, resulting in semantic distortion. Hierarchical Topic Models (HTMs) \cite{ncrp} are the first attempt for topic hierarchy, but ignore graph hierarchy, e.g., citations and hyperlinks.

Graph hierarchy is denoted by edge connectivity across documents, and topic hierarchy appears within text content of documents. 
Though some works, e.g., HGTM \cite{hgtm}, consider both hierarchies, they model both of them \emph{separately}, i.e., \emph{first} encoding graph hierarchy, \emph{then} learning topic hierarchy. Such ``cascaded'' method can not well integrate both hierarchies into unified representations, because topic hierarchy is neglected when encoding graph hierarchy. Topic hierarchy could reveal semantic similarity of documents and benefit graph hierarchical learning. Consequently, two hierarchies can not mutually enhance each other, and the representations are biased towards one hierarchy and neglect the other.


\textbf{Approach.} 
We propose GTFormer, a Hierarchical \textbf{G}raph \textbf{T}opic Modeling Trans\textbf{former}, integrating both topic hierarchy and graph hierarchy into a unified Transformer (Fig. \ref{fig:model}(a)). \emph{First}, to encode topic hierarchy, we design a \emph{topic tree} in latent semantic space and infer a \emph{hierarchical tree embedding} (Fig. \ref{fig:model}(b)). Documents with general content have high probability on root topic, while specific documents focus on leaf topics. \emph{Second}, to derive effective tree embedding, 
we design topic tree in hyperbolic space, which is more suitable than Euclidean space for hierarchical structure \cite{hgcn}. We design \emph{Hyperbolic Doubly Recurrent Neural Network}, modeling ancestral (parent-to-children) and fraternal (sibling-to-sibling) tree structure to recurrently derive hyperbolic tree embedding. (Fig. \ref{fig:model}(b-c)). In contrast, previous HTMs mainly operate in Euclidean space, leading to topic distortion. \emph{Third}, to deeply unify both topic and graph hierarchies, we insert both tree and graph representations into each Transformer layer. The contextualized modeling allows one to propagate information to the other, and the output representation integrates both hierarchies. 

\textbf{Contributions.} \emph{First}, we propose GTFormer to jointly model topic and graph hierarchies into a unified Transformer. To explore topic hierarchy, we design a topic tree and infer hierarchical tree representation. \emph{Second}, to better preserve both topic and graph hierarchies, we design in hyperbolic space and we propose Hyperbolic Doubly Recurrent Network. \emph{Third}, both hierarchies are unified into each Transformer layer for contextualized modeling. 
\section{Related Work}


\textbf{Topic models} are first designed with flat topics \cite{lda,nvdm,prodlda,bertopic,topicgpt,nguyen2024topic,wu2019short,wu2020learning,wu2020short,wu2021discovering,wu2022mitigating,wu2023infoctm,wu2024dynamic,wu2024thesis}. HTMs explore topic hierarchy, e.g., graphical \cite{ncrp,dhtg,wedtm,dirbn,pam,dpfa,gbn,hdp} and neural ones \cite{tsntm,htv,hntm,sawetm,traco,hyhtm}, but no one captures graph structure. Though Doubly RNN appears in \cite{drnn,tsntm,htv}, it is in Euclidean, not in hyperbolic space. Hyperbolic space has been shown to be more effective to capture hierarchy.

\textbf{Relational topic models} deal with graph-structured documents \cite{rtm,nrtm,adjacent_encoder,gtnn,lantm}. 
The recent HGTM \cite{hgtm} is the only one with both hierarchies, but is a cascaded method and is not effective to integrate both hierarchies. 

\textbf{Graph neural networks (GNNs)} are first proposed in Euclidean space \cite{gcn,gat,graphsage}. To model graph hierarchy, hyperbolic GNNs are proposed, e.g., HGNN \cite{hgnn}, HGCN \cite{hgcn}, HAT \cite{hat}, HTGN \cite{htgn}, $\kappa$GCN \cite{kgcn}. However, they mainly focus on graph structure, and do not deal with textual semantics.

\textbf{Text-attributed graph} combines GNNs and language models for both graph and text, e.g., GraphFormer \cite{graphformers}, Patton \cite{patton}, Heterformer \cite{heterformer}, Edgeformers \cite{edgeformers}, TAPE \cite{tape}, Specter \cite{specter}, LinkBERT \cite{linkbert}, etc. They consider both modalities, but no one models topic or graph \emph{hierarchy}. 
\section{Problem Formulation and Preliminaries}


$ \mathcal{G}=\{\mathcal{D},\mathcal{E}\} $ is a document graph. $ \mathcal{D}=\{d_i\}_{i=1}^N $ is a set of $ N $ documents. Each document $ d_i=\{w_{i,v}\}_{v=1}^{|d_i|}\subset\mathcal{V} $ is a sequence of words in vocabulary $ \mathcal{V} $. $ \mathcal{E}=\{e_{ij}\} $ is a set of edges. If there is an edge between documents $ i $ and $ j $, $ e_{ij}\in\mathcal{E} $. We follow \cite{hgtm} and model an undirected graph, $ e_{ij}=e_{ji} $, though our model is also applicable to directed graph. 
For document $ i $, its neighbor set $ \mathcal{N}(i) $ contains documents directly linked to $ i $.

Given $ \mathcal{G} $ as input, we propose a topic model that outputs unified document representations preserving topic hierarchy $ \mathcal{D} $ and graph hierarchy $ \mathcal{E} $. 
Appendix \ref{sec:notations} summarizes math notations.

\textbf{Hyperbolic geometry} is a non-Euclidean differential geometry with a constant negative curvature $ -1/K $ $ (K>0) $. Curvature measures how a geometric object deviates from a flat plane. 
We work with Hyperboloid model \cite{hyperboloid_model}, though our work is applicable to others, e.g., Poincar{\'e} ball \cite{poincare_model}.

\textbf{Hyperboloid model} is an $ n $-dimensional hyperbolic space $ \Bbb H^{n,K} $ where Minkowski self-inner product ($ \langle\cdot,\cdot\rangle_\mathcal{L} $) of its vectors is $ -K $,
\begin{equation}
\resizebox{0.88\columnwidth}{!}{
$ \begin{split}
    \Bbb H^{n,K}=&\{\textbf{x}\in\Bbb R^{n+1}|\langle\textbf{x},\textbf{x}\rangle_{\mathcal{L}}=-K,x_0>0\}\\
    \text{\quad where\quad}
    \langle\textbf{x}&,\textbf{y}\rangle_{\mathcal{L}}=-x_0y_0+x_1y_1+...+x_ny_n.
\end{split} $}
\end{equation}
For hyperbolic vector $ \textbf{x}\in\Bbb H^{n,K} $, the tangent space $ \mathcal{T}_{\textbf{x}}\Bbb H^{n,K} $ around $ \textbf{x} $ is first-order approximation of $ \Bbb H^{n,K} $ and is $ (n+1) $-dimensional Euclidean space.
\begin{equation}
     \mathcal{T}_{\textbf{x}}\Bbb H^{n,K}=\{\textbf{v}\in\Bbb R^{n+1}|\langle\textbf{x},\textbf{v}\rangle_{\mathcal{L}}=0\}.
\end{equation}

\textbf{Exponential and logarithmic maps.} The projection between hyperbolic and tangent space is achieved by exponential and logarithmic maps. For a hyperbolic vector $ \textbf{x}\in\Bbb H^{n,K} $ and one of its tangent vectors $ \textbf{v}\in\mathcal{T}_{\textbf{x}}\Bbb H^{n,K} $ $ (\textbf{v}\neq \textbf{0}) $, exponential map projects $ \textbf{v} $ to the hyperbolic space by
\begin{equation}
\resizebox{1\columnwidth}{!}{
$
\exp_{\textbf{x}}^K(\textbf{v})=\cosh\Big(\dfrac{||\textbf{v}||_{\mathcal{L}}}{\sqrt{K}}\Big)\textbf{x}+\sqrt{K}\sinh\Big(\dfrac{||\textbf{v}||_{\mathcal{L}}}{\sqrt{K}}\Big)\dfrac{\textbf{v}}{||\textbf{v}||_{\mathcal{L}}}. $
}
\end{equation}
$ ||\textbf{v}||_{\mathcal{L}}=\sqrt{\langle\textbf{v},\textbf{v}\rangle_{\mathcal{L}}} $ is the norm of $ \textbf{v}\in\mathcal{T}_{\textbf{x}}\Bbb H^{n,K} $. Reversely, for $ \textbf{x}\in\Bbb H^{n,K} $ and hyperbolic vector $ \textbf{y}\in\Bbb H^{n,K} $ $ (\textbf{x}\neq\textbf{y}) $, logarithmic map projects $ \textbf{y} $ to $ \textbf{x} $'s tangent space. $ d_{\mathcal{L}}^K(\textbf{x},\textbf{y}) $ is the distance between two hyperbolic vectors in Hyperboloid.
\begin{equation}
\label{eq:log_map}
\resizebox{0.88\columnwidth}{!}{
$ \begin{split}
    \log_{\textbf{x}}^K(\textbf{y})&=d_{\mathcal{L}}^K(\textbf{x},\textbf{y})\dfrac{\textbf{y}+\frac{1}{K}\langle\textbf{x},\textbf{y}\rangle_{\mathcal{L}}\textbf{x}}{||\textbf{y}+\frac{1}{K}\langle\textbf{x},\textbf{y}\rangle_{\mathcal{L}}\textbf{x}||_{\mathcal{L}}}\\
    \text{\quad where\quad}d_{\mathcal{L}}^K&(\textbf{x},\textbf{y})=\sqrt{K}\text{arcosh}(-\langle\textbf{x},\textbf{y}\rangle_{\mathcal{L}}/K).
\end{split} $}
\end{equation}

\textbf{Parallel transport.} For two hyperbolic vectors $ \textbf{x},\textbf{y}\in\Bbb H^{n,K} $ $ (\textbf{x}\neq\textbf{y}) $, parallel transport can transport $ \textbf{v}\in\mathcal{T}_{\textbf{x}}\Bbb H^{n,K} $ on $ \textbf{x} $’s tangent space to $ \textbf{y} $’s.
\begin{equation}
\resizebox{1\columnwidth}{!}{
$
    \text{PT}^K_{\textbf{x}\rightarrow\textbf{y}}(\textbf{v})=\textbf{v}-\dfrac{\langle\log_{\textbf{x}}^K(\textbf{y}),\textbf{v}\rangle_{\mathcal{L}}}{d_{\mathcal{L}}^K(\textbf{x},\textbf{y})^2}(\log_{\textbf{x}}^K(\textbf{y})+\log_{\textbf{y}}^K(\textbf{x})). $
    }
\end{equation}
\section{Model Architecture}

\begin{figure*}[t]
	\centering
	\includegraphics[width=0.9\linewidth]{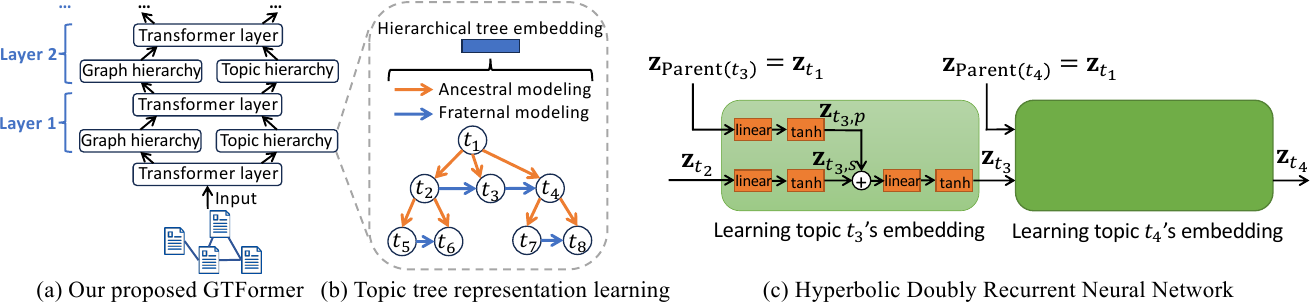}
	\vspace{-0.2cm}
	\caption{Illustration of (a) our proposed GTFormer, (b) topic tree embedding, and (c) Hyperbolic Doubly Recurrent Neural Network. Hyperbolic operations are omitted for clarity. Best seen in color.}
	\label{fig:model}
	\vspace{-0.4cm}
\end{figure*}

Fig. \ref{fig:model} shows the overall model with Hyperbolic Doubly Recurrent Network for hierarchical tree representation and a nested Transformer.

\subsection{Tree-structured Topic Hierarchy}
\label{sec:tree_structured_topic_hierarchy}

To preserve topic hierarchy, we construct a topic tree in latent semantic space, illustrated by Fig. \ref{fig:illustration}(b) and Fig. \ref{fig:model}(b). 
Documents with general concept present high probability on the root topic, while documents focusing on specific content sample topics near leaves. Thus, we calculate topic probability distribution of each document over topic tree, so as to differentiate semantic specificity of different documents and preserve topic hierarchy. Specifically, for each document $ i $, we calculate its path distribution $ \bm{\pi}_i $ and level distribution $ \bm{\delta}_i $, which are then combined to derive its topic distribution over topic tree $ \bm{\theta}_i\in\Delta^T $. $ \Delta^T $ is ($ T-1 $)-dimensional simplex. $ T $ is the number of topics on topic tree.

\textbf{Path distribution.} A path contains a sequence of topics, with increasing topic specificity from root to leaf. Different paths express different semantics. A document with certain themes present high probability on one path and low probabilities on others. In Fig. \ref{fig:illustration}(b), the left path shows reinforcement learning, while the right one is knowledge base. For path $ p $ with topics $ \{t_h\}_{h=1}^H $ where $ H $ is the path length, document $ i $'s path probability is
\begin{equation}
\label{eq:path_dist}
\resizebox{1\columnwidth}{!}{
$ \begin{split}
    \Pr(p=\{t_h\}_{h=1}^H)=&\Pr(t_H|t_{H-1})...\Pr(t_2|t_1)\Pr(t_1),
\end{split} $
}
\end{equation}
where $ \Pr(t_1)=\Pr(t_{\text{root}})=1 $. Topic $ t_h $ is one of the child topics of $ t_{h-1} $. Since a parent topic usually has more than one child, conditional probability $ \Pr(t_h|t_{h-1}) $ is the probability of selecting one of the child topics $ t_h $ given parent $ t_{h-1} $. We define it by tree-based stick-breaking construction:
\begin{equation}
\label{eq:path_dist_conditional_prob}
\resizebox{1\columnwidth}{!}{
$ \begin{split}
    \Pr(t_h|t_{h-1})=\sigma(t_h,i)&\prod_{t_h^\prime\in\text{LeftSibling}(t_h)}(1-\sigma(t_h^\prime,i)),
\end{split} $
}
\end{equation}
where $ h=1,2,...,H $. $ \sigma(t_h,i)\in[0,1] $ is the similarity between document $ i $ and topic $ t_h $, to be explained shortly. $ \text{LeftSibling}(t_h) $ contains left siblings of topic $ t_h $. Suppose a parent topic has three children ($ t_A $, $ t_B $, and $ t_C $), the probability of selecting each child topic is respectively $ \sigma(t_A,i) $, $ \sigma(t_B,i)(1-\sigma(t_A,i)) $, and $ (1-\sigma(t_A,i))(1-\sigma(t_B,i)) $, which are summed to one \cite{tsntm}. A document $ i $ with higher similarity to one child topic $ t_h $ than its siblings tends to have higher probability of selecting $ t_h $ on the path. Starting from the root topic, we repeat this selection process until we reach leaf topic, forming a path with Eq. \ref{eq:path_dist} as path probability. Finally, we calculate Eq. \ref{eq:path_dist} for every path and obtain path distribution $ \bm{\pi}_i=[\Pr(p_1),\Pr(p_2),...] $, which is unique to document $ i $, since $ \sigma(t_h,i) $ is document-specific.

\textbf{Hyperbolic doubly recurrent neural network.} We now explain similarity $ \sigma(t_h,i) $, which is calculated by topic $ t_h $'s and document $ i $'s embeddings, using Hyperbolic Doubly Recurrent Neural Network (HypDRNN). HypDRNN consists of two Hyperbolic Recurrent Neural Networks (HypRNNs) that respectively model the ancestral (parent-to-children) and fraternal (sibling-to-sibling) tree structure to preserve topic hierarchy. Specifically, a topic $ t $ has parent and siblings, thus we first use two HypRNNs to respectively calculate its ancestral hidden state $ \textbf{z}_{t,p} $ and fraternal hidden state $ \textbf{z}_{t,s} $, which are then combined to obtain $ t $'s hyperbolic hidden state $ \textbf{z}_t $. See Fig. \ref{fig:model}(b-c).

We first present topic $ t $'s ancestral (parent-to-children) hidden state. We have feature projection,
\begin{equation}
\label{eq:linear_transformation}
    \textbf{z}_{t,p}^\prime=\exp_{\textbf{0}}^K(\textbf{W}_p\log_{\textbf{0}}^K(\textbf{z}_{\text{Parent}(t)}))\in\Bbb H^{n,K}.
\end{equation}
$ \textbf{z}_{\text{Parent}(t)}\in\Bbb H^{n,K} $ is hyperbolic hidden state of $ t $'s parent. $ \textbf{W}_p\in\Bbb R^{(n+1)\times(n+1)} $ is Euclidean parameter. We project $ \textbf{z}_{\text{Parent}(t)} $ to tangent space for linear projection, whose result is mapped back to hyperbolic space. We use the origin $ \textbf{0}=[\sqrt{K},0,...,0]\in\Bbb H^{n,K} $ as projection reference point.

For bias addition, we use parallel transport. We initialize a Euclidean bias $ \textbf{b}_p\in\Bbb R^n $ and concatenate it with 0, i.e., $ \textbf{b}_p^\prime=[0||\textbf{b}_p]\in\mathcal{T}_{\textbf{0}}\Bbb H^{n,K} $. $ \textbf{b}_p^\prime $ is on origin's tangent space, due to $ \langle\textbf{b}_p^\prime,\textbf{0}\rangle_{\mathcal{L}}=0 $. We then transport $ \textbf{b}_p^\prime $ to the tangent space of $ \textbf{z}_{t,p}^\prime $, whose result is mapped back to hyperbolic space.
\begin{equation}
\label{eq:bias_addition}
    \textbf{z}_{t,p}=f_{\text{tanh}}^K\Big(\exp_{\textbf{z}_{t,p}^\prime}^K(\text{PT}^K_{\textbf{0}\rightarrow\textbf{z}_{t,p}^\prime}(\textbf{b}_p^\prime))\Big)\in\Bbb H^{n,K}.
\end{equation}
Here we use hyperbolic tanh activation $ f_{\text{tanh}}^K(\textbf{x})=\exp_{\textbf{0}}^K(\text{tanh}(\log_{\textbf{0}}^K(\textbf{x}))) $ and $ \text{tanh}(x)=\frac{e^x-e^{-x}}{e^x+e^{-x}} $. Summarizing Eqs. \ref{eq:linear_transformation}--\ref{eq:bias_addition}, we have ancestral hidden state $ \textbf{z}_{t,p}=f_{\text{HypRNN}}(\textbf{z}_{\text{Parent}(t)};\textbf{W}_p,\textbf{b}_p) $. Similarly, raternal hidden state is $ \textbf{z}_{t,s}=f_{\text{HypRNN}}(\textbf{z}_{\text{LeftSibling}(t)};\textbf{W}_s,\textbf{b}_s) $. Finally, we obtain topic $ t $'s hyperbolic hidden state $ \textbf{z}_t $ where $ \textbf{W}\in\Bbb R^{(n+1)\times(n+1)} $ is Euclidean parameter.
\begin{equation}
\label{eq:hyperbolic_hidden_state}
\resizebox{1\columnwidth}{!}{
    $ \textbf{z}_t=f^K_{\text{tanh}}\Big(\exp_{\textbf{0}}^K\Big(\textbf{W}\big(\log_{\textbf{0}}^K(\textbf{z}_{t,p})+\log_{\textbf{0}}^K(\textbf{z}_{t,s})\big)\Big)\Big)\in\Bbb H^{n,K}. $
    }
\end{equation}
Integrating Eqs. \ref{eq:linear_transformation}--\ref{eq:hyperbolic_hidden_state}, we have Hyperbolic Doubly RNN, $ \textbf{z}_t=f_{\text{HypDRNN}}(\textbf{z}_{t,p},\textbf{z}_{t,s}) $, which uses two Hyperbolic RNNs to model the information flow of ancestral and fraternal tree structure, hence the name of this module. See Fig. \ref{fig:model}(c). Hidden state $ \textbf{z}_t $ is also topic $ t $'s hyperbolic topic embedding.

Finally, the similarity $ \sigma(t,i) $ is calculated by topic $ t $'s and document $ i $'s hyperbolic embeddings, using Fermi-Dirac function $ \sigma(t,i)=[1+e^{d_{\mathcal{L}}^K(\textbf{z}_t,\textbf{d}_i^{(l)})^2}]^{-1} $. 
$ \textbf{d}_i^{(l)} $ is document $ i $'s hyperbolic embedding, to be explained shortly.

\textbf{Level distribution.} Having calculated path distribution, we discuss level distribution. A path contains a sequence of topics, each representing a level. Different levels represent different topic specificity. Root topic denotes the most general concept, while leaf topic focuses on specific sub-concept. 
Specifically, for a document $ i $, its probability at level $ h $ is another tree-based stick-breaking process,
\begin{equation}
    \delta_h=\sigma(h,i)\prod_{h^\prime=1}^{h-1}(1-\sigma(h^\prime,i)),
\end{equation}
where $ h=1,2,...,H $. $ \sigma(h,i)\in[0,1] $ is the similarity between level $ h $ and document $ i $. A general document presents high probability at root $ h=1 $, while a document with specific content falls on bottom $ h=H $. We calculate level similarity $ \sigma(h,i) $ by a separate Hyperbolic RNN, 
\begin{equation}
\begin{split}
    \textbf{z}_h=&f_{\text{HypRNN}}(\textbf{z}_{h-1};\textbf{W}_H,\textbf{b}_H)\in\Bbb H^{n,K},\\
    &\sigma(h,i)=[1+e^{d_{\mathcal{L}}^K(\textbf{z}_h,\textbf{d}_i)^2}]^{-1}.
\end{split}
\end{equation}
$ \textbf{z}_h\in\Bbb H^{n,K} $ is hyperbolic hidden state of level $ h $. $ f_{\text{HypRNN}}(\cdot) $ is a separate hyperbolic RNN with hyperbolic tanh activation (Eqs. \ref{eq:linear_transformation}--\ref{eq:bias_addition}). Level similarity $ \sigma(h,i) $ is Fermi-Dirac function. For document $ i $, we evaluate its probability of each level and obtain level distribution $ \bm{\delta}_i=[\delta_1,\delta_2,...,\delta_H] $.

Having obtained both path and level distributions, we now combine them to derive document $ i $'s topic distribution over topic tree. Specifically, given one path $ p=\{t_h\}_{h=1}^H $ and one level $ h $, we already narrow down to one topic $ t_h $ with probability $ \pi_i(p)\times\delta_i(h) $. Since there are multiple paths going through topic $ t_h $, the overall topic probability is $ \theta_{t_h}=\delta_i(h)\sum_{t_h\in p'}\pi_i(p^\prime) $, i.e., the summation of all paths having topic $ t_h $ at level $ h $. We repeat this process for every topic and obtain document $ i $'s topic distribution $ \bm{\theta}_i=[\theta_{i,t_1},\theta_{i,t_2},...,\theta_{i,T}] $ where $ T $ is the total number of topics on the tree. Documents with general content present high probability on the root topic $ \theta_{t_1} $, while documents focusing on specific content concentrate on leaf topics. Thus, this topic distribution $ \bm{\theta}_i $ preserves topic hierarchy.  

\textbf{Hyperbolic tree representation.} We use document $ i $'s hierarchical topic distribution $ \bm{\theta}_i $ and topic embeddings $ \{\textbf{z}_t\}_{t=1}^T $ to obtain \emph{hierarchical tree embedding} $ \textbf{e}_i\in\Bbb H^{n,K} $.
\begin{equation}
\label{eq:tree_embedding}
    \textbf{e}_i=\exp_{\textbf{0}}^K\Big(\sum_{t=1}^T \theta_{i,t}\log_{\textbf{0}}^K(\textbf{z}_t)\Big)\in\Bbb H^{n,K}.
\end{equation}
$ \textbf{z}_t\in\Bbb H^{n,K} $ is $ t $'s topic embedding. 
Tree embedding $ \textbf{e}_i $ preserves topic hierarchy and will later be inserted into Transformer. 

\subsection{Topic and Graph Joint Modeling}
\label{sec:tranformer}

We show hierarchical graph embedding and integrate both hierarchies into Transformer (Fig. \ref{fig:model}(a)).

\textbf{Hyperbolic graph representation.} We use Hyperbolic Graph Neural Network \cite{hgcn} to capture graph hierarchy. 
For document $ \textbf{d}_i^{(l)} $, we first linearly transform it by 
\begin{equation}
\label{eq:hgnn_linear_transformation}
\begin{split}
    \tilde{\textbf{d}}_i^{(l)\prime}=\exp_{\textbf{0}}^K(\textbf{W}_g\log_{\textbf{0}}^K(\textbf{d}_i^{(l)}))\in\Bbb H^{n,K}.
\end{split}
\end{equation}
We then evaluate its attention w.r.t. its neighbors and aggregate neighbor embeddings by Eq. \ref{eq:hgnn_neighbor_aggregation}. 
$ \text{softmax}(x)=\frac{e^x}{\sum_{x^\prime}e^{x^\prime}} $. Here $ \textbf{b}_{\text{att}}\in\Bbb R^{2(n+1)} $ is Euclidean parameter.
\begin{equation}
\label{eq:hgnn_neighbor_aggregation}
\resizebox{0.87\columnwidth}{!}{
$ \begin{split}
    &\alpha_{ij}=\text{softmax}\Big(\textbf{b}_{\text{att}}^\top[\log_{\textbf{0}}^K(\tilde{\textbf{d}}_i^{(l)}) \parallel \log_{\textbf{0}}^K(\tilde{\textbf{d}}_j^{(l)})]\Big),\\
    \textbf{g}_i^{(l)}&=\exp_{\textbf{0}}^K\Big(\dfrac{1}{2}\big(\log_{\textbf{0}}^K(\tilde{\textbf{d}}_i^{(l)})+\sum_{j \in \mathcal{N}(i)}\alpha_{ij}\log_{\textbf{0}}^K(\tilde{\textbf{d}}_j^{(l)})\big)\Big).
\end{split} $
}
\end{equation}
After HGNN, we obtain $ i $'s aggregated embedding $ \textbf{g}_i^{(l)}$ , preserving its hierarchical graph structure.

\textbf{Topic- and graph-nested Transformer encoding.} We have obtained both hierarchical tree embedding $ \textbf{e}_i^{(l)} $ at Eq. \ref{eq:tree_embedding} and hierarchical graph embedding $ \textbf{g}_i^{(l)} $ at Eq. \ref{eq:hgnn_neighbor_aggregation}. Both are calculated by document $ i $'s embedding $ \textbf{d}_i^{(l)} $ from the $ l $-th Transformer layer. Now we aim to insert them into the $ (l+1) $-th Transformer layer for hierarchical encoding. Specifically, we let $ \textbf{H}_i^{(l)}=[\textbf{H}_{i,\text{CLS}}^{(l)},\textbf{H}_{i,w_1}^{(l)},\textbf{H}_{i,w_2}^{(l)},...] $ denote the output from the $ l $-th Transformer layer. We concatenate $ \textbf{e}_i^{(l)} $ and $ \textbf{g}_i^{(l)} $ with $ \textbf{H}_i^{(l)} $, i.e., $ \tilde{\textbf{H}}_i^{(l)}=[\textbf{e}_i^{(l)}||\textbf{g}_i^{(l)}||\textbf{H}_i^{(l)}] $. After concatenation, $ \tilde{\textbf{H}}_i^{(l)} $ contains information of both hierarchical topic tree and hierarchical graph structure. To allow all the tokens in $ \tilde{\textbf{H}}_i^{(l)} $ to fully capture both hierarchies, we input it to the next Transformer layer for contextualized modeling.
\begin{equation}
    \textbf{H}_i^{(l+1)}=f_{\text{HypTRM}}(\tilde{\textbf{H}}_i^{(l)}).
\end{equation}
Here $ f_{\text{HypTRM}}(\cdot) $ is a Transformer layer in hyperbolic space. 
The building blocks of $ f_{\text{HypTRM}}(\cdot) $ are mostly the same as existing Transformer \cite{transformer}, except that the embeddings are hyperbolic and need to be projected between hyperbolic and tangent spaces. To make this paper self-contained, below we briefly present $ f_{\text{HypTRM}}(\cdot) $. We input $ \tilde{\textbf{H}}_i^{(l)} $ to the $ (l+1) $-th Transformer layer, where $ f_{\text{LN}}(\cdot) $ is layer normalization, and $ f_{\text{MLP}}(\cdot) $ is multi-layer perceptron \cite{transformer}.
\begin{equation}
\resizebox{0.88\columnwidth}{!}{
$ \begin{split}
    \tilde{\textbf{H}}_i^{(l)\prime}&=f_{\text{LN}}\Big(\log_{\textbf{0}}^K(\textbf{H}_i^{(l)})+\log_{\textbf{0}}^K(f_{\text{AsymMHA}}(\tilde{\textbf{H}}_i^{(l)}))\Big),\\
    &\textbf{H}_i^{(l+1)}=\exp_{\textbf{0}}^K\Big(f_{\text{LN}}(\tilde{\textbf{H}}_i^{(l)\prime}+f_{\text{MLP}}(\tilde{\textbf{H}}_i^{(l)\prime}))\Big).
\end{split} $
}
\end{equation}
We follow \cite{graphformers} and implement an \emph{asymmetric} multi-head attention $ f_{\text{AsymMHA}}(\cdot) $, where $ \textbf{K} $ and $ \textbf{V} $ are augmented with hierarchical topic and graph embeddings, while $ \textbf{Q} $ is not.
\begin{equation}
\resizebox{1\columnwidth}{!}{
$ \begin{split}
    f_{\text{AsyMHA}}(\tilde{\textbf{H}}_i^{(l)})=\text{softmax}&\Big(\dfrac{\log_{\textbf{0}}^K(\textbf{Q})\log_{\textbf{0}}^K(\textbf{K}^\top)}{\sqrt{n+1}}\Big)\log_{\textbf{0}}^K(\textbf{V}),\\
    \textbf{Q}=\exp_{\textbf{0}}^K(\log_{\textbf{0}}^K(\textbf{H}_i^{(l)})\textbf{W}&_Q^{(l)}),\;\  \textbf{K}=\exp_{\textbf{0}}^K(\log_{\textbf{0}}^K(\tilde{\textbf{H}}_i^{(l)})\textbf{W}_K^{(l)}),\;\  \\
    \textbf{V}=\exp_{\textbf{0}}^K&(\log_{\textbf{0}}^K(\tilde{\textbf{H}}_i^{(l)})\textbf{W}_V^{(l)}).
\end{split} $
}
\end{equation}
The output from the $ (l+1) $-th layer $ \textbf{H}_i^{(l+1)} $ preserves both hierarchies. 
We use [CLS] token to calculate hierarchical tree and graph embeddings for the current layer, which are concatenated and passed to the next layer. We repeat this \emph{layer-wise encoding} for $ L $ layers and obtain $ \textbf{d}_i=\textbf{H}_{i,\text{CLS}}^{(L)}\in\Bbb H^{n,K} $, document $ i $'s hierarchical embedding. 

\subsection{Training Objective}
\label{sec:training_objective}

\textbf{Decoding.} Since we preserve both topic and graph hierarchies, we present two decodings. For hierarchical topic decoding, we first use document $ i $'s final-layer embedding $ \textbf{d}_i $ to calculate its topic distribution $ \bm{\theta}_i $ by its path and level distributions. As in LDA, we then construct topic-word distribution for each topic $ t $ by 
$ \bm{\beta}_t=\text{softmax}(\textbf{U}\log_{\textbf{0}}^K(\textbf{z}_t))\in\Delta^{|\mathcal{V}|} $. Here $ \textbf{U}\in\Bbb R^{|\mathcal{V}|\times(n+1)} $ is a matrix of word embeddings, $ \mathcal{V} $ is vocabulary. 
The overall topic-word distribution is $ \bm{\beta}=[\bm{\beta}_1,...,\bm{\beta}_T]\in\Bbb R^{|\mathcal{V}|\times T} $. The reconstructed document is $ \hat{\textbf{d}}_i=\bm{\beta}\bm{\theta}_i\in\Delta^{|\mathcal{V}|} $. The topic modeling objective is $ \mathcal{L}_{\text{Topic}} $ where $ d_{i,w} $ is the word count of word $ w $ in document $ i $. For graph decoding, we use contrastive loss in Eq. \ref{eq:decoding}. 
\begin{equation}
\label{eq:decoding}
\resizebox{1\columnwidth}{!}{
$ \begin{split}
    \mathcal{L}&_{\text{Topic}}=-\sum_{w\in\mathcal{V}}d_{i,w}\log \hat{d}_{i,w},\\
    \mathcal{L}_{\text{Graph}}=-\log& \dfrac{e^{-d_{\mathcal{L}}^K(\textbf{d}_i,\textbf{d}_j)^2}}{e^{-d_{\mathcal{L}}^K(\textbf{d}_i,\textbf{d}_j)^2}+\sum_{j^\prime\in B}e^{-d_{\mathcal{L}}^K(\textbf{d}_i,\textbf{d}_{j^\prime})^2}}.
\end{split} $
}
\end{equation}
The overall loss is $ \mathcal{L}=\mathcal{L}_{\text{Graph}}+\lambda_{\text{Topic}}\mathcal{L}_{\text{Topic}} $. Hyperparameter $ \lambda_{\text{Topic}} $ controls the weight of $ \mathcal{L}_{\text{Topic}} $.

\textbf{Supervised version.} The above model is \emph{unsupervised}. If we also observe document labels, 
we design the \emph{supervised} version by adding a classifier $ \hat{\textbf{y}}_i= \text{softmax}(f_{\text{MLP}}(\log_{\textbf{0}}^K(\textbf{d}_i))) $. The supervised loss is $ \mathcal{L}_{\text{Sup}}=-\sum_{y^\prime}y_{i}^\prime\log\hat{y}_{i}^\prime $. The overall loss of the supervised version becomes $ \mathcal{L}=\mathcal{L}_{\text{Graph}}+\lambda_{\text{Topic}}\mathcal{L}_{\text{Topic}}+\lambda_{\text{Sup}}\mathcal{L}_{\text{Sup}} $.

\textbf{Continuously updating topic tree.} Different corpora contain documents of different topic hierarchy. To match the unique topic hierarchy of the given corpus, we update topic tree during training. For each topic $ t $, we calculate the proportion of words belonging to it by $ s_t=\frac{\sum_{d_i\in\mathcal{D}_{\text{train}}}|d_i|\theta_{i,t}}{\sum_{d_i\in\mathcal{D}_{\text{train}}}|d_i|} $ where $ |d_i| $ is the number of words in document $ i $. For a non-leaf topic whose $ s_t $ is greater than the adding threshold $ s_{\text{add}} $, we add a child topic, since it contains overly much semantics, and we split it into sub-concepts. Reversely, if the summation of topic $ t $ and its descendants $ \sum_{t^\prime\in\text{Des}(t)}s_{t^\prime} $ is smaller than the pruning threshold $ s_{\text{prune}} $, we remove topic $ t $ and its descendants, since they contain overly small proportion of semantics, and it is not necessary to keep them. Empirically, we set $ s_{\text{add}}=s_{\text{prune}}=0.05 $.  We summarize our model with Algorithm \ref{algo:training_algorithm}.
\section{Experiments}

\begin{table}
	\centering
	\caption{Dataset statistics.}
	\vspace{-0.2cm}
	\resizebox{0.75\columnwidth}{!}{
		\begin{tabular}{cccc}
			\toprule
			Name  & \#Documents & \#Links & \#Labels \\
			\midrule
                DS & 1,703 & 3,234 & 9 \\ 
			ML & 3,087 & 8,573 & 7 \\ 
			PL & 2,597 & 7,754 & 9 \\
                DBLP & 239,026 & 1,071,208 & N.A. \\
			COVID & 1,500 & 5,706 & 5 \\
			Web & 445,657 & 565,502 & N.A. \\
			\bottomrule
		\end{tabular}
	}
	\vspace{-0.5cm}
	\label{table:dataset_statistics}
\end{table}
\begin{table*}[!h]
	\centering
	\caption{Unsupervised document classification with Micro F1 and Macro F1 scores (in percentage).}
	\vspace{-0.3cm}
	\resizebox{\linewidth}{!}{
		\begin{tabular}{c|c|rrrr|rrrr}
			\toprule
			\multirow{2}{*}{Category} & \multirow{2}{*}{Model} & \multicolumn{4}{c|}{Micro F1} & \multicolumn{4}{c}{Macro F1} \\
			\cline{3-10}
			{} & {} & DS & ML & PL & COVID & DS & ML & PL & COVID \\
			\hline
			\multirow{4}{*}{\shortstack{\textbf{Flat topic models}}} & ProdLDA & 51.4$ \pm $1.1 & 65.3$ \pm $1.0 & 49.8$ \pm $2.5 & 72.7$ \pm $1.7 & 40.1$ \pm $4.3 & 67.4$ \pm $1.4 & 48.4$ \pm $1.8 & 73.3$ \pm $1.7 \\
			{} & ETM & 42.2$ \pm $2.4 & 46.2$ \pm $1.2 & 39.8$ \pm $0.8 & 67.2$ \pm $1.8 & 31.1$ \pm $3.4 & 42.8$ \pm $1.6 & 32.1$ \pm $1.2 & 67.4$ \pm $1.7 \\
			{} & GATON & 39.7$ \pm $1.3 & 61.1$ \pm $0.7 & 51.2$ \pm $1.0 & 70.5$ \pm $1.2 & 29.9$ \pm $1.8 & 58.2$ \pm $0.8 & 44.6$ \pm $1.5 & 70.7$ \pm $1.1 \\
            {} & GNTM & 37.0$ \pm $1.3 & 60.2$ \pm $3.1 & 50.0$ \pm $1.9 & 73.2$ \pm $2.2 & 26.4$ \pm $2.1 & 56.4$ \pm $3.4 & 43.0$ \pm $1.6 & 73.2$ \pm $2.1 \\
			\hline
			\multirow{5}{*}{\shortstack{\textbf{Hierarchical topic models}}} & nCRP & 27.5$ \pm $3.3 & 28.6$ \pm $1.7 & 25.2$ \pm $2.5 & 41.7$ \pm $4.4 & 16.7$ \pm $2.2 & 21.6$ \pm $1.8 & 16.7$ \pm $2.5 & 41.5$ \pm $4.5 \\
			{} & TSNTM & 39.5$ \pm $3.7 & 62.0$ \pm $1.0 & 47.8$ \pm $1.3 & 74.1$ \pm $3.2 & 28.9$ \pm $4.1 & 58.8$ \pm $1.0 & 38.9$ \pm $1.6 & 73.9$ \pm $3.2 \\
			{} & HTV & 29.7$ \pm $2.1 & 37.3$ \pm $4.2 & 29.2$ \pm $5.4 & 61.6$ \pm $4.3 & 13.7$ \pm $2.2 & 32.0$ \pm $4.1 & 21.3$ \pm $4.7 & 61.9$ \pm $4.7 \\
            {} & HyperMiner & 41.3$ \pm $1.3 & 53.4$ \pm $0.1 & 45.3$ \pm $0.4 & 50.2$ \pm $4.5 & 30.1$ \pm $1.9 & 43.9$ \pm $2.5 & 38.6$ \pm $1.4 & 48.6$ \pm $2.9 \\
            {} & TraCo & 46.7$ \pm $2.1 & 64.4$ \pm $4.2 & 47.5$ \pm $2.0 & 60.4$ \pm $3.8 & 36.7$ \pm $1.9 & 61.2$ \pm $5.3 & 38.6$ \pm $2.1 & 61.3$ \pm $3.4 \\
			\hline
			\multirow{4}{*}{\shortstack{\textbf{Topic models for document graph}}} & AdjEnc & 58.8$ \pm $1.2 & 72.5$ \pm $1.1 & 61.2$ \pm $1.0 & 74.8$ \pm $2.4 & 54.6$ \pm $1.5 & 68.3$ \pm $1.0 & 49.3$ \pm $0.6 & 69.8$ \pm $2.3 \\
			{} & LANTM & 56.8$ \pm $2.4 & 72.2$ \pm $0.7 & 61.7$ \pm $1.1 & 80.3$ \pm $1.7 & 54.7$ \pm $0.8 & 68.6$ \pm $1.0 & 54.6$ \pm $1.2 & 80.2$ \pm $1.7 \\
			{} & GTNN & 52.9$ \pm $1.4 & 68.1$ \pm $0.7 & 58.8$ \pm $1.2 & 70.9$ \pm $1.0 & 42.8$ \pm $3.3 & 64.7$ \pm $1.3 & 52.4$ \pm $1.3 & 70.8$ \pm $0.9 \\
            {} & HGTM & 65.6$ \pm $1.5 & 82.1$ \pm $0.9 & 68.3$ \pm $0.5 & 81.6$ \pm $0.5 & 62.3$ \pm $1.6 & 80.1$ \pm $0.9 & 63.6$ \pm $0.7 & 81.2$ \pm $0.6 \\
            \hline
			\multirow{5}{*}{\shortstack{\textbf{Text-attributed graph models}}} & BERT+HGCN & 62.5$ \pm $1.4 & 78.9$ \pm $0.9 & 62.7$ \pm $1.5 & 75.5$ \pm $1.3 & 58.6$ \pm $1.3 & 77.0$ \pm $1.1 & 57.0$ \pm $1.3 & 75.1$ \pm $1.6 \\
            {} & LLaMA2+HGCN & 67.3$ \pm $1.6 & 83.0$ \pm $0.5 & 65.8$ \pm $1.1 & 81.3$ \pm $0.3 & 64.4$ \pm $1.2 & 80.9$ \pm $0.6 & 59.4$ \pm $1.9 & 81.0$ \pm $0.6 \\
            {} & Specter & 63.1$ \pm $0.1 & 77.2$ \pm $0.8 & 63.7$ \pm $1.6 & 80.3$ \pm $1.5 & 59.5$ \pm $1.2 & 75.2$ \pm $0.8 & 59.3$ \pm $1.9 & 80.0$ \pm $1.7 \\
            {} & LinkBERT & 47.5$ \pm $2.2 & 61.5$ \pm $3.2 & 47.9$ \pm $0.4 & 72.0$ \pm $1.2 & 40.0$ \pm $2.7 & 59.9$ \pm $3.8 & 39.6$ \pm $0.1 & 76.1$ \pm $0.6 \\
            {} & Patton & 65.1$ \pm $1.8 & 82.4$ \pm $1.0 & 70.0$ \pm $1.6 & 78.6$ \pm $1.5 & 60.0$ \pm $2.4 & 80.4$ \pm $1.1 & 65.5$ \pm $1.9 & 77.9$ \pm $1.7 \\
			\hline
			\multirow{1}{*}{\shortstack{\textbf{Hyperbolic graph transformer}}} & FPS-T & 60.9$ \pm $2.7 & 74.7$ \pm $2.6 & 67.2$ \pm $3.9 & 76.0$ \pm $2.4 & 53.4$ \pm $4.7 & 73.4$ \pm $3.0 & 63.0$ \pm $3.2 & 75.7$ \pm $2.4 \\
			\hline
			{\textbf{Our proposed model} (\emph{unsupervised})} & GTFormer & \textbf{69.1}$ \pm $\textbf{0.6} & \textbf{84.5}$ \pm $\textbf{0.9} & \textbf{70.9}$ \pm $\textbf{2.4} & \textbf{82.3}$ \pm $\textbf{1.1} & \textbf{65.8}$ \pm $\textbf{0.8} & \textbf{82.8}$ \pm $\textbf{1.0} & \textbf{65.8}$ \pm $\textbf{2.1} & \textbf{82.2}$ \pm $\textbf{1.1} \\
			\bottomrule
		\end{tabular}
	}
	\vspace{-0.2cm}
	\label{table:unsupervised_classification}
\end{table*}
\begin{table*}[!h]
	\centering
	\caption{Supervised document classification with Micro F1 and Macro F1 scores (in percentage).}
	\vspace{-0.3cm}
	\resizebox{\linewidth}{!}{
		\begin{tabular}{c|c|rrrr|rrrr}
			\toprule
			\multirow{2}{*}{Category} & \multirow{2}{*}{Model} & \multicolumn{4}{c|}{Micro F1} & \multicolumn{4}{c}{Macro F1} \\
			\cline{3-10}
			{} & {} & DS & ML & PL & COVID & DS & ML & PL & COVID \\
			\hline
			\multirow{3}{*}{\shortstack{\textbf{\emph{Supervised} version}}} & TSNTM & 54.9$ \pm $2.1 & 72.8$ \pm $1.5 & 63.3$ \pm $0.5 & 84.1$ \pm $1.3 & 50.8$ \pm $2.4 & 68.6$ \pm $1.3 & 56.1$ \pm $0.8 & 84.0$ \pm $1.2 \\
            {} & HGTM & 68.2$ \pm $0.8 & 83.8$ \pm $0.5 & 72.2$ \pm $1.4 & 86.3$ \pm $1.7 & 63.9$ \pm $1.5 & 82.6$ \pm $0.7 & 67.4$ \pm $2.0 & \textbf{86.2}$ \pm $\textbf{1.9} \\
            {} & Patton & 67.8$ \pm $3.5 & 84.1$ \pm $2.4 & 73.5$ \pm $0.5 & 81.5$ \pm $1.2 & 62.9$ \pm $3.0 & 83.2$ \pm $2.3 & 69.5$ \pm $1.7 & 80.8$ \pm $1.6 \\
			\hline
            \multirow{4}{*}{\shortstack{\textbf{Text classification models}}} & TextGCN & 66.8$ \pm $1.0 & 78.3$ \pm $0.7 & 67.5$ \pm $0.7 & 83.7$ \pm $0.5 & 61.6$ \pm $0.4 & 76.0$ \pm $0.8 & 61.4$ \pm $1.1 & 79.6$ \pm $0.5 \\
			{} & HyperGAT & 70.2$ \pm $0.4 & 80.0$ \pm $0.4 & 65.8$ \pm $2.5 & 84.3$ \pm $1.2 & 65.4 $ \pm $0.9 & 78.9$ \pm $0.5 & 60.2$ \pm $2.5 & 81.3$ \pm $0.8 \\
			{} & HINT & 45.7$ \pm $3.5 & 69.5$ \pm $1.1 & 55.4$ \pm $2.3 & 85.7$ \pm $1.5 & 42.1$ \pm $2.6 & 64.8$ \pm $3.9 & 44.3$ \pm $3.2 & 85.8$ \pm $1.5 \\
            {} & G2P2 & 58.0$ \pm $2.2 & 71.7$ \pm $3.2 & 65.3$ \pm $0.4 & 77.5$ \pm $1.2 & 51.1$ \pm $2.7 & 68.7$ \pm $3.8 & 61.1$ \pm $0.1 & 75.6$ \pm $0.6 \\
			\hline
			{\textbf{Our proposed model} (\emph{supervised})} & GTFormer & \textbf{72.2}$ \pm $\textbf{1.1} & \textbf{86.5}$ \pm $\textbf{0.5} & \textbf{74.5}$ \pm $\textbf{1.1} & \textbf{85.3}$ \pm $\textbf{1.0} & \textbf{69.3}$ \pm $\textbf{2.0} & \textbf{85.1}$ \pm $\textbf{0.7} & \textbf{71.4}$ \pm $\textbf{1.0} & \textbf{86.2}$ \pm $\textbf{1.1} \\
			\bottomrule
		\end{tabular}
	}
	\vspace{-0.2cm}
	\label{table:supervised_classification}
\end{table*}
\begin{table*}[!h]
	\centering
	\caption{Topic coherence NPMI (left, in percentage) and perplexity (right).}
	\vspace{-0.3cm}
	\resizebox{\linewidth}{!}{
		\begin{tabular}{c|rrrrrr|rrrrrr}
			\toprule
			\multirow{2}{*}{Model} & \multicolumn{6}{c|}{Topic Coherence NPMI (higher is better)} & \multicolumn{6}{c}{Perplexity (lower is better)} \\
			\cline{2-13}
			{}  & DS & ML & PL & COVID & DBLP & Web & DS & ML & PL & COVID & DBLP & Web \\
			\hline
			ProdLDA & 10.5$ \pm $0.3 & 10.9$ \pm $0.7 & 12.1$ \pm $0.7 & 12.0$ \pm $0.7 & 9.9$ \pm $0.6 & 21.2$ \pm $0.2 & 7.97$ \pm $0.00 & 7.99$ \pm $0.00 & 7.92$ \pm $0.00 & 7.82$ \pm $0.00 & 8.18$ \pm $0.00 & 8.34$ \pm $0.00 \\
			ETM & 7.3$ \pm $0.2 & 7.1$ \pm $0.2 & 8.7$ \pm $0.1 & 8.2$ \pm $0.7 & 9.5$ \pm $0.5 & 16.4$ \pm $0.6 & 7.92$ \pm $0.00 & 7.96$ \pm $0.00 & 7.94$ \pm $0.00 & 7.80$ \pm $0.00 & 8.66$ \pm $0.00 & 8.52$ \pm $0.00 \\
			GATON & 12.2$ \pm $0.2 & 17.4$ \pm $1.0 & 5.4$ \pm $1.1 & 13.8$ \pm $1.2 & 7.2$ \pm $0.8 & 4.8$ \pm $1.1 & 8.83$ \pm $0.07 & 8.37$ \pm $0.02 & 8.38$ \pm $0.03 & 8.42$ \pm $0.00 & 8.35$ \pm $0.00 & 8.33$ \pm $0.00 \\
			GNTM & 11.6$ \pm $0.5 & 12.1$ \pm $0.3 & 15.4$ \pm $0.7 & 13.8$ \pm $0.8 & 15.2$ \pm $0.2 & 23.8$ \pm $0.3 & 7.18$ \pm $0.01 & 6.91$ \pm $0.01 & 6.83$ \pm $0.01 & 7.69$ \pm $0.01 & 7.52$ \pm $0.00 & 7.79$ \pm $0.00 \\
			\hline
			nCRP & 2.6$ \pm $0.4 & 2.2$ \pm $0.1 & 2.2$ \pm $0.1 & 3.0$ \pm $0.1 & 2.8$ \pm $0.3 & 2.8$ \pm $0.0 & 6.91$ \pm $0.05 & 6.94$ \pm $0.02 & 6.87$ \pm $0.02 & 7.69$ \pm $0.05 & 8.00$ \pm $0.02 & 7.71$ \pm $0.04 \\
			TSNTM & 11.5$ \pm $0.9 & 12.1$ \pm $0.6 & 15.1$ \pm $0.8 & 14.1$ \pm $0.8 & 15.1$ \pm $1.0 & \textbf{26.6}$ \pm $\textbf{2.3} & 7.75$ \pm $0.02 & 6.92$ \pm $0.01 & 6.83$ \pm $0.01 & 7.64$ \pm $0.04 & 7.68$ \pm $0.01 & \textbf{7.35}$ \pm $\textbf{0.03} \\
			HTV & 11.2$ \pm $1.2 & 10.8$ \pm $1.0 & 13.3$ \pm $1.8 & 16.6$ \pm $2.5 & 12.1$ \pm $0.6 & 26.5$ \pm $0.9 & 7.78$ \pm $0.03 & 6.95$ \pm $0.02 & 6.83$ \pm $0.03 & 7.62$ \pm $0.04 & 7.53$ \pm $0.01 & 7.44$ \pm $0.01 \\
            HyperMiner & 12.9$ \pm $0.2 & 15.9$ \pm $0.2 & 20.0$ \pm $1.5 & 9.9$ \pm $0.8 & 17.3$ \pm $0.3 & 15.3$ \pm $0.5 & 7.86$ \pm $0.23 & 7.73$ \pm $0.19 & 7.69$ \pm $0.20 & 8.04$ \pm $0.22 & 9.65$ \pm $0.02 & 8.54$ \pm $0.01 \\
            TraCo & 11.0$ \pm $0.3 & 11.5$ \pm $0.2 & 11.2$ \pm $0.3 & 12.7$ \pm $0.5 & 15.4$ \pm $0.4 & 18.6$ \pm $0.2 & 7.83$ \pm $0.08 & 7.69$ \pm $0.01 & 7.65$ \pm $0.01 & 8.04$ \pm $0.02 & 8.64$ \pm $0.00 & 7.67$ \pm $0.01 \\
			\hline
			AdjEnc & 12.0$ \pm $0.2 & 9.9$ \pm $0.9 & 11.3$ \pm $0.9 & 13.8$ \pm $0.4 & 9.2$ \pm $0.2 & 15.2$ \pm $0.1 & 8.06$ \pm $0.02 & 7.65$ \pm $0.05 & 7.62$ \pm $0.04 & 6.96$ \pm $0.00 & 8.71$ \pm $0.02 & 8.26$ \pm $0.01 \\
			LANTM & 6.4$ \pm $0.5 & 5.4$ \pm $0.3 & 7.2$ \pm $0.8 & 8.6$ \pm $0.3 & N.A. & N.A. & 7.58$ \pm $0.03 & 8.63$ \pm $0.00 & 8.48$ \pm $0.00 & 8.48$ \pm $0.00 & N.A. & N.A. \\
			GTNN & 9.9$ \pm $1.5 & 7.2$ \pm $0.6 & 5.8$ \pm $0.6 & 13.5$ \pm $2.7 & 8.3$ \pm $0.5 & 7.9$ \pm $1.6 & 7.77$ \pm $0.04 & 7.75$ \pm $0.02 & 7.73$ \pm $0.01 & 7.96$ \pm $0.00 & 9.39$ \pm $0.01 & 8.26$ \pm $0.01 \\
            HGTM & 17.1$ \pm $1.4 & 19.0$ \pm $2.6 & 21.9$ \pm $2.8 & 23.3$ \pm $3.1 & 18.5$ \pm $1.2 & 25.0$ \pm $1.7 & 7.46$ \pm $0.03 & 6.89$ \pm $0.02 & 6.81$ \pm $0.00 & 7.60$ \pm $0.01 & 7.77$ \pm $0.02 & 7.71$ \pm $0.01 \\
			\hline
			HINT & 9.3$ \pm $1.3 & 6.6$ \pm $2.2 & 8.6$ \pm $2.4 & 11.6$ \pm $3.0 & N.A. & N.A. & 8.04$ \pm $0.07 & 8.45$ \pm $0.08 & 8.51$ \pm $0.28 & 8.84$ \pm $0.12 & N.A. & N.A. \\
			\hline
			GTFormer & \textbf{19.1}$ \pm $\textbf{1.2} & \textbf{20.6}$ \pm $\textbf{0.4} & \textbf{23.2}$ \pm $\textbf{1.5} & \textbf{24.0}$ \pm $\textbf{1.3} & \textbf{20.2}$ \pm $\textbf{1.0} & 26.2$ \pm $0.8 & \textbf{7.40}$ \pm $\textbf{0.03} & \textbf{6.82}$ \pm $\textbf{0.04} & \textbf{6.68}$ \pm $\textbf{0.04} & \textbf{6.79}$ \pm $\textbf{0.03} & \textbf{7.49}$ \pm $\textbf{0.00} & 7.58$ \pm $0.00 \\
			\bottomrule
		\end{tabular}
	}
	\vspace{-0.4cm}
	\label{table:topic_analysis}
\end{table*}

\begin{table*}[t]
	\centering
	\caption{Link prediction with AUC score (in percentage).}
	\vspace{-0.3cm}
	\resizebox{\linewidth}{!}{
		\begin{tabular}{c|c|rrrrrr}
			\toprule
			\multirow{2}{*}{Category} & \multirow{2}{*}{Model} & \multicolumn{6}{c}{Link prediction AUC} \\
			\cline{3-8}
			{} & {} & DS & ML & PL & COVID & DBLP & Web \\
			\hline
			\multirow{4}{*}{\shortstack{\textbf{Flat topic models}}} & ProdLDA & 76.8$ \pm $0.5 & 82.5$ \pm $0.2 & 78.9$ \pm $0.4 & 80.1$ \pm $0.8 & 92.6$ \pm $0.1 & 82.4$ \pm $0.0 \\
            {} & ETM & 71.0$ \pm $1.3 & 70.3$ \pm $1.6 & 72.5$ \pm $1.1 & 87.2$ \pm $0.9 & 70.3$ \pm $1.2 & 79.4$ \pm $0.0 \\
            {} & GATON & 76.0$ \pm $0.6 & 75.9$ \pm $1.5 & 64.5$ \pm $0.5 & 70.2$ \pm $1.2 & 82.6$ \pm $0.8 & 87.6$ \pm $0.1 \\
            {} & GNTM & 77.0$ \pm $0.9 & 79.3$ \pm $0.8 & 73.2$ \pm $0.3 & 76.8$ \pm $0.4 & 93.7$ \pm $0.0 & 86.3$ \pm $0.0 \\
            \hline
            \multirow{5}{*}{\shortstack{\textbf{Hierarchical topic models}}} & nCRP & 62.2$ \pm $3.3 & 58.0$ \pm $0.9 & 60.1$ \pm $3.1 & 70.8$ \pm $0.8 & 56.3$ \pm $1.1 & 57.2$ \pm $0.0 \\
            {} & TSNTM & 72.7$ \pm $1.9 & 77.8$ \pm $1.5 & 75.5$ \pm $0.9 & 70.8$ \pm $0.8 & 84.6$ \pm $0.8 & 87.4$ \pm $0.8 \\
            {} & HTV & 66.2$ \pm $2.5 & 69.9$ \pm $1.9 & 68.3$ \pm $4.6 & 86.0$ \pm $2.0 & 84.9$ \pm $1.0 & 86.1$ \pm $0.8 \\
            {} & HyperMiner & 64.0$ \pm $1.0 & 63.7$ \pm $1.3 & 62.5$ \pm $2.6 & 71.5$ \pm $2.0 & 75.1$ \pm $2.8 & 78.2$ \pm $1.3 \\
            {} & TraCo & 72.5$ \pm $0.5 & 75.0$ \pm $3.1 & 70.0$ \pm $1.6 & 76.4$ \pm $0.9 & 87.3$ \pm $1.2 & 87.5$ \pm $1.1 \\
            \hline
            \multirow{4}{*}{\shortstack{\textbf{Topic models for document graph}\\(LANTM cannot run on large data DBLP and Web)}} & AdjEnc & 81.7$ \pm $0.4 & 81.0$ \pm $1.1 & 80.8$ \pm $1.7 & 79.8$ \pm $0.6 & 95.5$ \pm $0.0 & 73.2$ \pm $0.0 \\
            {} & LANTM & 78.4$ \pm $0.6 & 78.7$ \pm $0.7 & 82.2$ \pm $0.3 & 93.6$ \pm $0.3 & N.A. & N.A. \\
            {} & GTNN & 71.5$ \pm $1.1 & 76.6$ \pm $0.9 & 73.7$ \pm $1.2 & 84.3$ \pm $1.0 & 95.3$ \pm $0.4 & 74.3$ \pm $0.2 \\
            {} & HGTM & 94.5$ \pm $0.3 & 89.9$ \pm $0.8 & 91.3$ \pm $0.3 & \textbf{95.7}$ \pm $\textbf{0.2} & 96.6$ \pm $0.3 & \textbf{91.3}$ \pm $\textbf{0.1} \\
            \hline
            \multirow{5}{*}{\shortstack{\textbf{Text-attributed graph models}}} & BERT+HGCN & 93.6$ \pm $0.4 & 91.9$ \pm $0.7 & 91.0$ \pm $0.8 & 91.5$ \pm $0.2 & 97.5$ \pm $0.1 & 90.2$ \pm $0.0 \\
            {} & LLaMA2+HGCN & 94.2$ \pm $0.2 & 92.5$ \pm $0.5 & 92.2$ \pm $0.8 & 92.2$ \pm $0.3 & 97.1$ \pm $0.0 & 91.5$ \pm $0.0 \\
            {} & Specter & 90.0$ \pm $1.3 & 88.1$ \pm $0.8 & 87.6$ \pm $0.8 & 87.2$ \pm $1.0 & 97.1$ \pm $0.2 & 86.3$ \pm $0.2 \\
            {} & LinkBERT & 66.9$ \pm $2.1 & 71.2$ \pm $1.5 & 66.3$ \pm $0.7 & 74.9$ \pm $1.2 & 75.3$ \pm $0.8 & 70.4$ \pm $0.5 \\
            {} & Patton & 93.7$ \pm $1.6 & 92.1$ \pm $0.7 & 92.2$ \pm $0.5 & 91.9$ \pm $1.3 & 97.6$ \pm $0.1 & 87.5$ \pm $0.2 \\
            \hline
            \multirow{1}{*}{\shortstack{\textbf{Hyperbolic transformer} (cannot run on large data)}} & FPS-T & 92.4$ \pm $2.1 & 90.1$ \pm $1.8 & 91.6$ \pm $0.7 & 89.1$ \pm $1.6 & N.A. & N.A. \\
            \hline
            \multirow{4}{*}{\shortstack{\textbf{Text classification models}\\(cannot run on DBLP and Web with no labels)}} & TextGCN & 83.2$ \pm $0.4 & 76.5$ \pm $0.5 & 68.2$ \pm $0.4 & 87.1$ \pm $0.4 & N.A. & N.A. \\
            {} & HyperGAT & 84.5$ \pm $0.2 & 82.0$ \pm $0.8 & 77.5$ \pm $1.0 & 87.1$ \pm $0.4 & N.A. & N.A. \\
			{} & HINT & 72.6$ \pm $2.1 & 71.7$ \pm $1.4 & 69.7$ \pm $1.4 & 86.6$ \pm $0.2 & N.A. & N.A. \\
            {} & G2P2 & 84.5$ \pm $0.7 & 82.9$ \pm $1.0 & 83.3$ \pm $0.1 & 85.0$ \pm $1.3 & N.A. & N.A. \\
            \hline
            {\textbf{Our proposed model}} & GTFormer & \textbf{95.6}$ \pm $\textbf{0.4} & \textbf{93.4}$ \pm $\textbf{0.5} & \textbf{93.4}$ \pm $\textbf{0.6} & 93.2$ \pm $0.7 & \textbf{98.1}$ \pm $\textbf{0.2} & \textbf{91.4}$ \pm $\textbf{0.1} \\
			\bottomrule
		\end{tabular}
	}
	\vspace{-0.2cm}
	\label{table:link_prediction}
\end{table*}

\begin{figure*}
	\centering
	\includegraphics[width=1\linewidth]{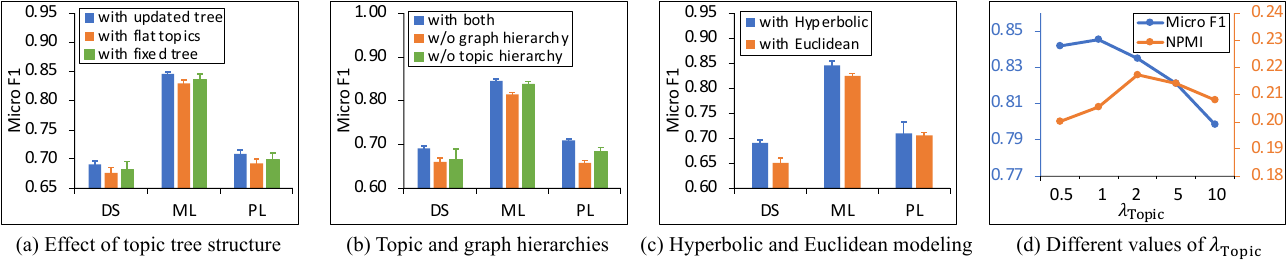}
	\vspace{-0.6cm}
	\caption{Ablation analysis of our model. Best seen in color.}
	\label{fig:model_analysis}
	\vspace{-0.42cm}
\end{figure*}


\textbf{Datasets.} We use six datasets in Table \ref{table:dataset_statistics}. Cora \cite{cora} is a citation network with abstracts as texts and citations as links. We follow \cite{cora2} and create three independent datasets, \textbf{DS}, \textbf{ML}, and \textbf{PL}. \textbf{DBLP} \cite{aminer} is another citation graph. \textbf{COVID} \cite{hgtm} is a Coronavirus news corpus with hyperlinks among articles. \textbf{Web} \cite{web} is a Webpage hyperlink network. Appendix \ref{sec:dataset_preprocessing} contains dataset preprocessing \cite{wu2024topmost}.

\textbf{Baselines.} Since GTFormer is topic model, we mainly compare to topic models. \emph{i}) \textbf{Flat topic models} do not learn any hierarchy, ProdLDA \cite{prodlda}, ETM \cite{etm}, GATON \cite{gaton}, GNTM \cite{gntm}. \emph{ii}) \textbf{Hierarchical topic models}, nCRP \cite{ncrp}, TSNTM \cite{tsntm}, HTV \cite{htv}, HyperMiner \cite{hyperminer}, TraCo \cite{traco}. \emph{iii}) \textbf{Topic models for document graph} capture text and graph, AdjEnc \cite{adjacent_encoder}, LANTM \cite{lantm}, GTNN \cite{gtnn}, HGTM \cite{hgtm}. HGTM has topic and graph hierarchies in a \emph{cascaded} method. \emph{iv}) \textbf{Text-attributed graph models.} 
Strictly speaking, they are not topic models, nor baselines. For completeness, we still compare to BERT+HGCN, LLaMA2+HGCN \cite{llama2}, Specter \cite{specter}, LinkBERT \cite{linkbert}, Patton \cite{patton}. For BERT+HGCN, we first use BERT to encode each document, then apply HGCN, in a \emph{cascaded} method. 
\emph{v}) \textbf{Hyperbolic graph transformer} is designed in hyperbolic space. It is not a topic model, either. We still compare to FPS-T \cite{fps_t}. \emph{vi}) \textbf{Text classification} has label supervision, TextGCN \cite{textgcn}, HyperGAT \cite{hypergat}, HINT \cite{hint}, G2P2 \cite{g2p2}. HINT has a topic tree. Besides, we also convert the best unsupervised baselines to their supervised version (TSNTM, HGTM, Patton).

We set $ \tau=10 $, $ \lambda_{\text{Topic}}=1 $. For the supervised model, $ \lambda_{\text{Sup}}=1 $. 
We initialize topic tree with three levels, each non-leaf topic has three children. We initialize model with scientific parameters \cite{scibert} for DS, ML, PL, DBLP, and general ones \cite{bert} for others. Experiments are done on an NVIDIA A40 GPU. Each result is obtained by 5 runs. 
Appendix \ref{sec:implementation_environment} has more details.

\subsection{Quantitative Evaluation}

\textbf{Document classification.} We use document classification for evaluation, as in LDA. We use 80\% documents and links within them for training (of which 10\% are for validation). We infer test document embeddings and classify them. 

\emph{Unsupervised training} does not involve labels and $ \lambda_{\text{Sup}}=0 $. We 
use $ \kappa $NN ($ \kappa=5 $) as classifier. We report Micro and Macro F1 scores at Table \ref{table:unsupervised_classification}.

\emph{Supervised training} uses labels for training 
We convert the best baselines to the supervised version by adding classifier $ f_{\text{MLP}}(\cdot) $. Table \ref{table:supervised_classification} shows results.

HGTM achieves better accuracy than other topic models, due to its modeling of both hierarchies. Our model is better than HGTM, since we capture both hierarchies in a nested manner and better integrate them into representations.


\textbf{Topic coherence.} 
To evaluate if keywords of each topic coherently reflect the same meaning, we follow ProdLDA and use Google Web 1T 5-gram Version 1 \cite{google_5gram} as external corpus to evaluate their NPMI score. Only topic models produce topic interpretability and can be evaluated, thus others are excluded. Table \ref{table:topic_analysis}(left) shows that 
HGTM has higher results than other baselines, due to its topic and graph hierarchies. We unify both hierarchies in a nested way and improve NPMI.

\textbf{Perplexity.} As in LDA, we report perplexity of test documents. Since perplexity, $ e^{-\frac{\log \Pr(\mathcal{D}_{\text{test}})}{\sum_{d\in\mathcal{D}_{\text{test}}}|d|}} $, varies much w.r.t. its power, we report its power at Table \ref{table:topic_analysis}. Lower is better. We model both hierarchies to differentiate documents, thus outperform baselines, except on Web where TSNTM is slightly better, because the text is more informative than its graph. But we are still better than it on other tasks.

\textbf{Link prediction.} 
We predict links within 20\% test documents. We 
evaluate the probability of a link for Euclidean models by $ \Pr(e_{ij})\propto e^{-||\textbf{d}_i-\textbf{d}_j||^2}$, and for hyperbolic models by $ \Pr(e_{ij})\propto e^{-d_{\mathcal{L}}^K(\textbf{d}_i,\textbf{d}_j)^2}$. 
We report AUC at Table \ref{table:link_prediction}. 
Among baselines, HGTM predicts links more accurately than others, since it models both topic and graph hierarchies. We achieve better results than it, due to nested modeling. 

See Appendix \ref{sec:qualitative_topic_analysis} for topic visualization.

\subsection{Model Analysis}

\textbf{Topic tree structure.} We design two settings: \emph{i}) replacing hierarchical tree with flat topics; \emph{ii}) fixing the tree structure during learning. Fig. \ref{fig:model_analysis}(a) shows that a flat structure decreases the result, since a hierarchical tree captures topic hierarchy to improve the result. A fixed tree decreases the result, since the pre-defined tree is not suitable for the corpus. 

\textbf{Topic and graph hierarchies.} 
We respectively remove each hierarchy from the model in Fig. \ref{fig:model_analysis}(b). The model with both hierarchies performs the best. The ablated models 
drop the result. This observation verifies that both hierarchies are useful. 

\textbf{Hyperbolic modeling.} 
We replace all hyperbolic operations with their Euclidean ones in Fig. \ref{fig:model_analysis}(c). Hyperbolic space is helpful to better preserve hierarchy and improve result than Euclidean space.

\textbf{Different $ \lambda_{\text{Topic}} $.} We vary $ \lambda_{\text{Topic}} $ in Fig. \ref{fig:model_analysis}(d). For classification, an appropriate value maintains result, while a high value hurts result, since a high value influences graph loss. For topic coherence, gradually increasing $ \lambda_{\text{Topic}} $ improves NPMI, 
while a high value hurts the result. Taking the balance, we set $ \lambda_{\text{Topic}}=1 $.

\section{Conclusion}

We design GTFormer, a topic model with topic and graph hierarchies. We design Hyperbolic Doubly RNN for tree embedding. Topic and graph hierarchies are inserted in Transformer. 

\clearpage
\section*{Limitations}

Here we identify two limitations in terms of training corpus and language.

\textbf{Training corpus.} We assume the corpus is truthful. If it has fake news, they may appear in the topics, causing potential negative impact. To mitigate it, we could use fake news detection model to remove fake documents, then apply our model on the remaining truthful documents.

\textbf{Language.} Corpora used in this paper mainly consist of English vocabulary. The language models are also mostly trained on English-based corpora. Since our model is corpus-agnostic, extracting multilingual information may be feasible with an appropriate corpus.
\section*{Ethics Statement}

We do not foresee any undesired implications stemming from our work. Conversely, we hope that our work can advance AI Ethics research.

\bibliography{custom}

\appendix
\section{Summary of Mathematical Notations}
\label{sec:notations}

We summarize the main mathematical notations used in the main paper in Table \ref{table:notations}.

\begin{table*}[h]
	\centering
	\caption{Summary of main mathematical notations.}
		\begin{tabular}{c|l}
			\toprule
			Notation  & Description  \\
			\hline
            $ \mathcal{G} $ & a document graph \\
			$ \mathcal{D} $ & a corpus of documents, $ \mathcal{D}=\{d_i\}_{i=1}^{N} $ \\
			$ N $ & number of documents in the corpus, $ N=|\mathcal{D}| $ \\
			$ d_i $ & document $ i $ containing a sequence of words, $ d_i=\{w_{i,v}\}_{v=1}^{|d_i|}\subset\mathcal{V} $ \\
			$ \mathcal{V} $ & vocabulary \\
			$ |d_i| $ & number of words in document $ i $ \\
			$ \mathcal{E} $ & a set of graph edges connecting documents, $ \mathcal{E}=\{e_{ij}\} $ \\
			$ \mathcal{N}(i) $ & the neighbor set of document $ i $ \\
            $ \Bbb H^{n,K} $ & Hyperboloid model with dimension $ n $ and curvature $ -1/K $ \\
			$ \mathcal{T}_{\textbf{x}}\Bbb H^{n,K} $ & tangent (Euclidean) space around hyperbolic vector $ x\in\Bbb H^{n,K} $ \\
			$ \exp_{\textbf{x}}^K(\textbf{v}) $ & exponential map, projecting tangent vector $ \textbf{v} $ to hyperbolic space \\
			$ \log_{\textbf{x}}^K(\textbf{y}) $ & logarithmic map, projecting hyperbolic vector $ \textbf{y} $ to $ \textbf{x} $'s tangent space \\
			$ d_{\mathcal{L}}^K(\textbf{x},\textbf{y}) $ & hyperbolic distance between hyperbolic vectors $ \textbf{x} $ and $ \textbf{y} $ \\
			$ \text{PT}_{\textbf{x}\rightarrow\textbf{y}}^K(\textbf{v}) $ & parallel transport, transporting $ \textbf{v} $ from $ \textbf{x} $'s tangent space to $ \textbf{y} $'s \\
			$ H $ & length of a path on topic tree \\
            $ \sigma(t,i) $ & similarity between topic $ t $ and document $ i $ \\
            $ \bm{\pi}_i $ & path distribution of document $ i $ over topic tree \\
            $ \textbf{z}_{t,p} $ & hyperbolic ancestral hidden state of topic $ t $ \\
            $ \textbf{z}_{t,s} $ & hyperbolic fraternal hidden state of topic $ t $ \\
			$ \textbf{z}_t $ & hyperbolic hidden state of topic $ t $ \\
            $ \sigma(h,i) $ & similarity between topic $ t $ and document $ i $ \\
            $ \textbf{z}_h $ & hyperbolic hidden state of level $ h $ \\
			$ \bm{\delta}_i $ & level distribution of document $ i $ over topic tree \\
            $ \bm{\theta}_i $ & topic distribution of document $ i $ over topic tree \\
            $ \textbf{e}_i $ & hierarchical tree embedding of document $ i $ \\
            $ T $ & number of topics on topic tree \\
            $ \textbf{g}_i $ & hierarchical graph embedding of document $ i $ \\
			$ \textbf{U} $ & a matrix of word embeddings, $ \textbf{U}\in\Bbb R^{|\mathcal{V}|\times(n+1)} $ \\
			$ \bm{\beta} $ & topic-word distribution $ \bm{\beta}\in\Bbb R^{T\times |\mathcal{V}|} $ \\
			\bottomrule
		\end{tabular}
	\label{table:notations}
\end{table*}
\section{Learning Algorithm}
\label{sec:algorithm}

We summarize the learning process of our model and formulate it in Algorithm \ref{algo:training_algorithm}. Code and datasets are submitted as supplementary materials. We will release them upon publication.

\begin{algorithm}
	\caption{Training Process of GTFormer}
	\label{algo:training_algorithm}
	\begin{flushleft}
		\hspace*{\algorithmicindent}\textbf{Input}: A document graph $ \mathcal{G} $ with documents $ \mathcal{D} $ and graph edges $ \mathcal{E} $, a predefined topic tree structure, hyperparameters $ \lambda_{\text{HTM}} $, $ \tau $, and $ \lambda_{\text{sup}} $. \\
		\hspace*{\algorithmicindent}\textbf{Output}: Document embeddings $ \{\textbf{h}_i\}_{i=1}^N $ and topic-word distribution $ \bm{\beta} $.
	\end{flushleft} 
	\begin{algorithmic}[1] 
		\State Initialize model parameters.
		\While{not converged}
        \For{document $ i\in\mathcal{D} $}
        \State Feature initialization 
        and obtain hyperbolic embedding $ \textbf{H}^{(l=0)}_i $.
        \Statex \quad\qquad\textit{ // Get the initial hyperbolic token embeddings}
        \State Pass to one hyperbolic Transformer layer and obtain $ \textbf{H}^{(l=1)}_i=f_{\text{HypTRM}}(\textbf{H}^{(l=0)}_i) $.
		\Statex \quad\qquad\textit{ // $ (L-1) $ topic- and graph-Transformer layers}
		\For{$ l=2,...,L $}
		\State Obtain document embedding $ \textbf{d}_i^{(l)}=\textbf{H}^{(l)}_{i,\text{CLS}} $. 
        \State Use Hyperbolic Doubly RNN to obtain hyperbolic topic embeddings $ \{\textbf{z}_t\}_{t=1}^T $.
        \State Evaluate path distribution $ \bm{\pi}_i^{(l)} $ and level distribution $ \bm{\delta}_i^{(l)} $ using document embedding $ \textbf{d}_i^{(l)} $ and topic embeddings $ \{\textbf{z}_t\}_{t=1}^T $.
        \Statex \qquad\qquad\textit{ // Hierarchical tree embedding}
        \State Obtain hierarchical tree embedding $ \textbf{e}_i^{(l)} $. 
        \Statex \qquad\qquad\textit{ // Capture graph hierarchy by HGNNs}
        \State Obtain hierarchical graph embedding $ \textbf{g}_i^{(l)} $. 
        \Statex \qquad\qquad\textit{ // Insert both hierarchies into Transformer}
        \State Obtain the output from the $ l $-th hyperbolic Transformer layer $ \textbf{H}_i^{(l+1)}=f_{\text{HypTRM}}([\textbf{e}_i^{(l)}||\textbf{g}_i^{(l)}||\textbf{H}_i^{(l)}]). $
		\EndFor
        \EndFor
        \Statex \quad\textit{ // Optimization}
		\State Minimize objective function $ \mathcal{L} $ with Adam optimizer.
		\Statex \quad\textit{ // Update topic tree structure}
		\State Update topic tree based on adding threshold $ \lambda_{\text{add}} $ and pruning threshold $ \lambda_{\text{prune}} $.
		\EndWhile
	\end{algorithmic}
\end{algorithm}

\section{Dataset Preprocessing}
\label{sec:dataset_preprocessing}

Here we introduce dataset preprocessing. We will release code and datasets upon publication.

Cora \cite{cora} is a paper citation graph where each document is an academic paper with abstract, and each graph edge is a citation between two documents. We follow \cite{cora2} and created three independent datasets, Data Structure (\textbf{DS}), Machine Learning (\textbf{ML}), and Programming Language (\textbf{PL}).

\textbf{DBLP} \cite{aminer} is anther academic paper citation graph. We used \textit{DBLP-Citation-network V4} version\footnote{\url{https://www.aminer.org/citation}}. We removed documents with no words and documents with no citations. After removal, we obtain 239,026 documents and 1,071,208 citation links.

\textbf{COVID} is a Coronavirus news corpus available online\footnote{\url{https://aylien.com/coronavirus-news-dataset/}}, collected from multiple publishers. Each document is a news article and has a category for the content of the article. We selected five categories, \emph{economy, business, and finance}, \emph{education}, \emph{health}, \emph{labour}, and \emph{sports}. For each category, we randomly selected 300 news articles, resulting in a corpus of 1,500 articles in total. Since we did not observe graph edges connecting these articles, we compared documents' $ tf-idf $ similarity and induced edges by $ \kappa $NN ($ \kappa=5 $), resulting in 5,706 links in total.

\textbf{Web} is a Webpage hyperlink graph publicly available online\footnote{\url{https://snap.stanford.edu/data/memetracker9.html}}. Each Webpage is a news article containing the most frequent phrases and quotes. Each page has hyperlinks to other related pages. The Webpages in this dataset were published between August 2008 through April 2009. We collected Webpages published between August 2008 through December 2008. For each Webpage, we used Breadth First Search algorithm to collect its neighbors. We remove pages with less than 30 words, resulting in 445,657 documents and 565,505 hyperlinks in total. We did not observe any ground-truth categories of these documents.
\section{Implementation Environment}
\label{sec:implementation_environment}

Here we introduce the detailed implementation details and environment for reproducibility purpose. For our model, we choose hyperparameters based on the performance on validation set (Document classification task in the main paper explains how we split validation set). The results in the main paper are obtain by 5 independent runs. The standard deviations reported in the main paper are 1-sigma error bars and are obtained by calling its corresponding function in Excel library. All the experiments were done on Linux server with an NVIDIA A40 GPU with 46,068 MiB. Its operating system is CentOS Linux 7 (Core). We implemented our proposed model GTFormer using Python 3.10 as programming language and PyTorch 2.0.0 as deep learning library. Other frameworks include NumPy 1.23.1, sklearn 0.23.2, and scipy 1.5.2. We emphasize that the main focus of our model is effectiveness, instead of running efficiency. But for completeness, we still make a short comment on execution time. Our model is efficient, on the largest dataset Web, the training takes less than 40 hours to converge. We will release code and datasets upon publication.
\section{Qualitative Topic Analysis}
\label{sec:qualitative_topic_analysis}

In the main paper, we mainly present quantitative topic analysis, including topic coherence and perplexity results. Here we further provide qualitative topic analysis as a case study.

\begin{figure*}[h]
	\centering
	\includegraphics[width=0.8\linewidth]{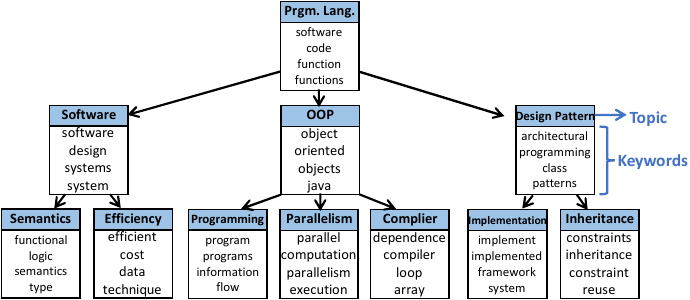}
	\caption{Topic tree structure learned on PL dataset.}
	\label{fig:topic_interpretability_pl}
\end{figure*}

\textbf{Topic interpretability.} To intuitively understand what topic tree structure our model learns and what keywords each topic contains, here we plot topic tree and keywords of each topic on PL dataset at Fig. \ref{fig:topic_interpretability_pl}. Here we show top-4 keywords of each topic for clarity purpose. For each topic, we manually summarize its keywords into one word or phrase. For topic hierarchy of other datasets, our submitted code can produce topic hierarchy after convergence for every dataset.

Overall, the learned topic tree has three levels. The root topic \textit{Programming Language} is split into three concepts at the second level, \textit{Software Analysis}, \textit{Object Oriented Programming (OOP)}, and \textit{Design Pattern}. For Software Analysis topic, the corpus seems to contain documents about \textit{Semantics} of programming language and \textit{Efficiency} of the program. Similarly, for Object Oriented Programming topic, papers in this corpus mainly talk about three sub-concepts, \textit{Programming}, \textit{Parallelism}, and \textit{Compiler}, all of which are related to OOP. Similar topic hierarchy can also be observed on the Design Pattern topic, which is split into \textit{Implementation} and \textit{Inheritance} topics.

\begin{figure*}[h]
	\centering
	\includegraphics[width=1\linewidth]{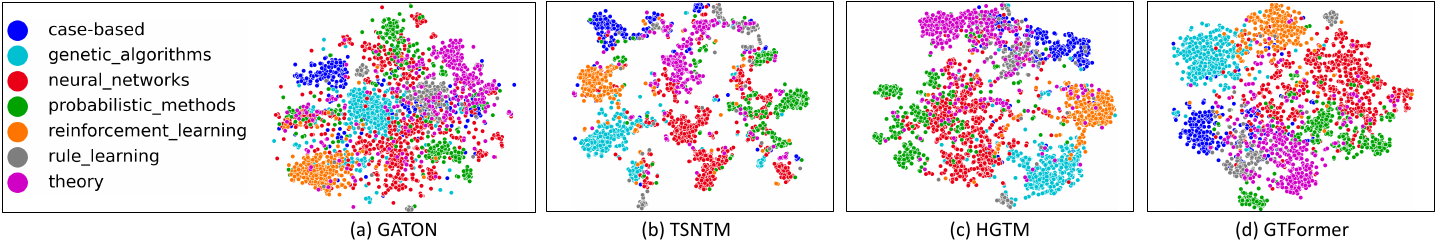}
	\caption{Visualization on ML dataset.}
	\label{fig:visualization}
\end{figure*}

\textbf{Topic visualization.} 
We use t-SNE \cite{tsne} to project document embeddings into 2D space and color embeddings using documents' labels in Fig. \ref{fig:visualization}. Since our model is a topic model, we mainly select representative topic models for visualization. GATON does not incorporate topic hierarchy or graph hierarchy, thereby its document embeddings of different categories tend to mix together. By modeling topic hiearchy, TSNTM produces clearer separation among different categories. HGTM and our model capture both topic hierarchy and graph hierarchy, and produce similar separation based on visual observation.

\end{document}